\documentclass[conference]{IEEEconf}
\IEEEoverridecommandlockouts
\usepackage{cite}
\usepackage{amsmath,amssymb,amsfonts}
\usepackage{optidef}
\usepackage{graphicx}
\usepackage{multirow}
\usepackage{textcomp}
\usepackage{xcolor}
\def\BibTeX{{\rm B\kern-.05em{\sc i\kern-.025em b}\kern-.08em
    T\kern-.1667em\lower.7ex\hbox{E}\kern-.125emX}}
    
\usepackage{mathtools}
\usepackage{wasysym}
\usepackage{bm}
\usepackage{bbm}

\usepackage[font=footnotesize]{subcaption}
\usepackage{booktabs}
\usepackage{array}
\usepackage{dblfloatfix}
\newcolumntype{L}[1]{>{\raggedright\let\newline\\\arraybackslash\hspace{0pt}}m{#1}}
\newcolumntype{C}[1]{>{\centering\let\newline\\\arraybackslash\hspace{0pt}}m{#1}}
\newcolumntype{R}[1]{>{\raggedleft\let\newline\\\arraybackslash\hspace{0pt}}m{#1}}

\usepackage{balance}
\usepackage{graphicx}

\usepackage{multirow}
\usepackage{tabularx}

\usepackage{algpseudocode}
\usepackage[linesnumbered,ruled,vlined,noend]{algorithm2e}
\SetKwRepeat{Do}{do}{while}

\usepackage{hyperref}
\usepackage{xcolor}
\hypersetup{
    colorlinks,
    linkcolor={red!70!black},
    citecolor={blue!70!black},
    urlcolor={blue!90!black}
}
\usepackage{comment}

\makeatletter
\let\oldabs\abs
\def\abs{\@ifstar{\oldabs}{\oldabs*}}
\let\oldnorm\norm
\def\norm{\@ifstar{\oldnorm}{\oldnorm*}}
\makeatother

\title{\LARGE \bf Haptic Shoulder for Rendering Biomechanically Accurate Joint Limits for Human-Robot Physical Interactions}

\author{Elizabeth Peiros$^{\dagger,1}$, Calvin Joyce$^{\dagger,2}$, Tarun Murugesan$^{\dagger,2}$, Roger Nguyen$^2$, Isabella Fiorini$^2$, Rizzi Galibut$^2$,\\Michael C. Yip$^1$, \IEEEmembership{Senior Member, IEEE}%
\thanks{$^\dagger$ Equal Contribution. This work was supported by the U.S. Army's Telemedicine and Advanced Technology Research Center under Project W81XWH-22-C-0089.}%
\thanks{$^1$ E. Peiros and M.C. Yip are with the Electrical and Computer Engineering Department, University of California, San Diego, La Jolla, CA 92093 USA. {\tt\footnotesize \{epeiros, yip\}@ucsd.edu}}%
\thanks{$^2$ C. Joyce, T. Murugesan, R. Nguyen, I. Fiorini, and R. Galibut are with the Mechanical and Aerospace Engineering Department, University of California, San Diego, La Jolla, CA 92093 USA. {\tt\footnotesize \{cajoyce, tmuruges, rqnguyen, ifiorini, rgalibut\}@ucsd.edu}}%
}

\begin{document}

\maketitle
\thispagestyle{empty}
\pagestyle{empty}

\begin{abstract}
Human-robot physical interaction (pHRI) is a rapidly evolving research field with significant implications for physical therapy, search and rescue, and telemedicine. However, a major challenge lies in accurately understanding human constraints and safety in human-robot physical experiments without an IRB and physical human experiments. Concerns regarding human studies include safety concerns, repeatability, and scalability of the number and diversity of participants. This paper examines whether a physical approximation can serve as a stand-in for human subjects to enhance robot autonomy for physical assistance.  This paper introduces the SHULDRD (Shoulder Haptic Universal Limb Dynamic Repositioning Device), an economical and anatomically similar device designed for real-time testing and deployment of pHRI planning tasks onto robots in the real world. SHULDRD replicates human shoulder motion, providing crucial force feedback and safety data. The device's open-source CAD and software facilitate easy construction and use, ensuring broad accessibility for researchers. By providing a flexible platform able to emulate infinite human subjects, ensure repeatable trials, and provide quantitative metrics to assess the effectiveness of the robotic intervention, SHULDRD aims to improve the safety and efficacy of human-robot physical interactions.
\end{abstract}


\section{Introduction}

\vspace{-1mm} 

Numerous tasks including in-home assistive care, telemedicine, casualty extraction, and physical therapy could benefit greatly from the addition of physical robotic assistance, specifically assistance with tasks involving moving the human body. Works have shown promise for assisting with dressing \cite{wang2023one} by holding and moving clothes and weight shifting \cite{1241924} by tugging on bed sheets as a proxy to grasping and lifting the person. While these tasks push robotic assistance forward they are missing human contact due to the challenge of conducting human trials and safety concerns. 

In medical diagnostics, there have been research and commercial implementations of impedance and admittance controllers that allow the robot to follow the contour of a patient with a model for the safety of human soft tissue. In research, there has been work to teleoperate and automate heart listening \cite{10089533} and automate arthritis diagnosis with ultrasound \cite{frederiksen2022ultrasound}. Similarly, these controllers are used for human-robot collaboration physical therapy tasks\cite{tejwanidemonstrating}. However, there is still a lack of a method for achieving dexterous manipulation in interactions with people. Being able to grasp, tug, and maneuver the human body would dramatically increase the type of assistance robots could provide as shown in \cite{peiros2023finding} for search and rescue tasks.

An area in which this type of contact is currently prevalent is wearable assistance technology. Many are cleverly designed to align with and move with the wearer's natural anatomy utilizing exoskeleton technology and advanced high-fidelity human modeling to optimize design. However, human intervention does not provide adequate protection. Studies show bone fractures and skin and soft tissue damage from lower limb exoskeletons are a risk of device use\cite{he2017risk}. The sim-to-real gap of high-fidelity human models and the high variance across real humans can lead to these injuries. This lesson from wearable technology begets the necessity for real-world online learning and higher consideration for user variation in addition to close human models.



\begin{figure}[t]
\centerline{\includegraphics[width=\linewidth]{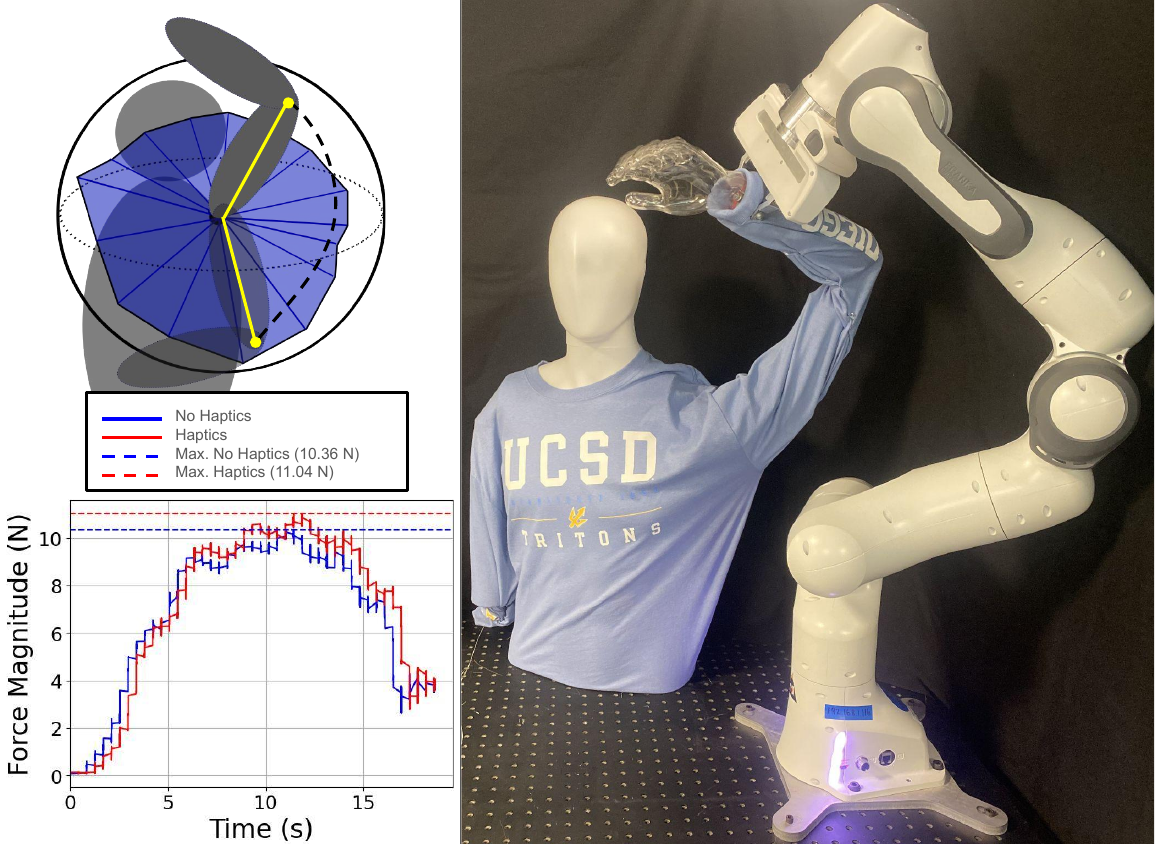}}
\caption{SHULDRD used in a human-robot physical interaction experiment where SHULDRD simulates a human shoulder. The image on the upper left shows the shoulder position relative to its complex spherical joint limits. The graph on the lower left shows the activation of the actuators rendering the elastic forces mimicking those produced by human tendons when the arm goes beyond its reachable space.}
\label{fig: results}
\vspace{-1mm}
\end{figure}


As the previous example shows, robots must first properly understand safety to achieve their full potential in these tasks: end ranges of motion in the joints, soft tissue damage\cite{zhao2024high}, and dislocations in the complex joint spaces of the human body. An ideal physical model would render enough information and variable cases to ensure robust and safe robotic trajectories without the risk to a human subject.





The next roadblock is the means to evaluate this safety. There are designs on the market for robotic shoulders that mimic or are inspired by human anatomy. Humanoids are one example and some manipulators could be considered shoulder-like with their similar range of motion achieved through redundant degrees of freedom. For human replacement in pHRI, neither humanoids nor serial link robotic manipulators are ideal. They are designed for actuation, not rendering virtual environments, and are costly if damaged. 

To bridge this gap, a simple and economical haptic shoulder is proposed for testing human-robot physical interaction planning tasks. The device is designed to replicate the anatomical range of motion of the human shoulder joint so that human participants do not need to be used as testing and development subjects. An extra key benefit will be providing force feedback, customizable individual user parameters, and data unavailable from a human subject which can be used to enhance learning algorithms and safety. The CAD is open source and simplified for construction by researchers with non-mechanical backgrounds. The materials used are chosen for flexibility and low cost. The software is open source and includes the models for shoulder complex reach cone joint limits.



\section{Related works}

Three bodies of work correspond to this task of recreating human limb motion (1) models of bio-mechanically accurate shoulders, (2) robotic platforms designed to move people, and (3) physical designs for human-like robotic shoulders.

Advanced human models are simulated in software like OpenSim \cite{delp2007opensim} and AnyBody modeling system\cite{damsgaard2006analysis}. These models are state-of-the-art for developing robotic platforms. However, simulated environments are missing the stochasticity of the real world, in real-time. 

While most wearable and rehabilitation technology designs align with anatomy, some designs have deviated from this technique and are used to move people safely. One example technique is whole-arm manipulation (WAM). Researchers have taken approaches to mimic human-like whole-arm manipulation and integrated compliance. Researchers have utilized reinforcement learning to produce capabilities to lift large baskets and simple human models \cite{yuan2019reinforcement, goncalves2022punyo}.

\begin{table}[t]
\vspace{2mm}
\centering
\caption{Comparison of Feasible Platforms} 
\label{tab:path_length_run_time}
\resizebox{0.49\textwidth}{!}{%
\footnotesize
\begin{tabular}{clccc}
    \toprule
    \multicolumn{2}{c}{Requirement Number} & Frank Panda & “Kojiro” Humanoid Shoulder & Ours \\
    \midrule
     1& Single Joint Center & $\chi$ & $\checkmark$ & $\checkmark$ \\
     2& ROM & $\checkmark$ & $\chi$ & $\checkmark$  \\
     3& Back Drivable Actuators & $\chi$ & $\chi$ & $\checkmark$  \\
     4& Accurate Measurements & $\checkmark$ & $\chi$ & $\checkmark$\\
     5& Safety & $\checkmark$ & $\chi$ & $\checkmark$ \\
    \bottomrule
    \end{tabular}
    \vspace{-1mm}
}
\end{table}

Shoulder replication mechanisms are typically categorized as serially linked or bio-mimetic designs. Examples of serially linked shoulders include humanoids \cite{grebenstein2011dlr,iwata2009design,diftler2011robonaut,5649567,7354208}, which offer sufficient force and range of motion but lack backdrivability and a singular joint center. A human-inspired design using tendon-driven ball-and-socket joints is seen in \cite{4651221}. While closely mimicking human kinematics, it is limited in humeral rotation and range of motion. Other mechanical designs with a high range of motion such as \cite{10.1115/1.4035802} and \cite{9415699}, reach a significant range, but still fall short of the full human shoulder range.




\section{Methods}

To make a useful device for pHRI there are essential characteristics the device must include to replicate the most important aspects of human anatomy critical for the safety of the human, real-world path planning experiments, and the safety of the robotic system being used. These requirements are listed in Table \ref{tab: requirements}.

The final system, referred to as SHULDRD (Shoulder Haptic Universal Limb Dynamic Repositioning Device), is presented with a clear analysis of how it compares to the desired capabilities.

\begin{table}[t]
\vspace{2mm}
\centering
\caption{Requirements for Haptic Shoulder Platform} 
\label{tab: requirements}
\resizebox{0.49\textwidth}{!}{%
\footnotesize
\begin{tabular}{clcc}
    \toprule
     num & \multicolumn{2}{c}{Requirement}\\
    \midrule
    1 & \multicolumn{2}{c}{singular joint center (no series linkage designs)}\\
    \midrule
    2 & \multicolumn{2}{c}{maximize reachable space of shoulder to ensure coverage of entire human shoulder range of motion} \\
    \midrule
    3 & \multicolumn{2}{c}{ability to give force feedback in 3D space of the shoulder joint}  \\
    \midrule
    4 & \multicolumn{2}{c}{measures the position of the shoulder with greater accuracy than traditional vision systems} \\
    \midrule
    5 & \multicolumn{2}{c}{ensure forces rendered are safe for operators and robotic arms that may be used in experiments} \\
    \bottomrule
    \end{tabular}
    }
    \vspace{-1mm}
\end{table}


\subsection{Possible Kinematic Chains for Mimicry}
\label{sec: kinematics}

\begin{figure}[t]
\centerline{\includegraphics[width=80mm]{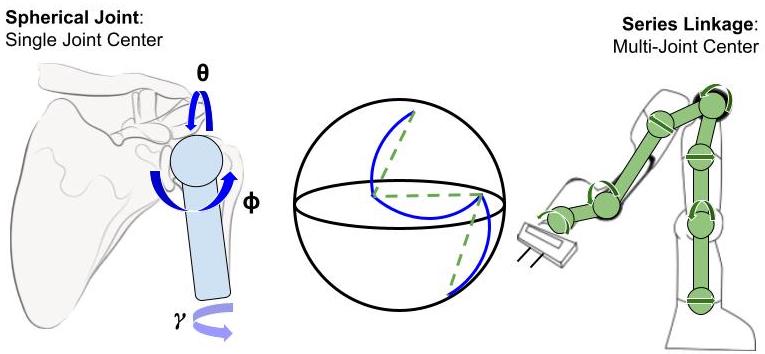}}
\caption{The image to the left shows an exact match for the kinematics of the shoulder joint, where $\theta$ and $\phi$ represent the flexion/extension (forward / backward) and abduction/adduction (out to the side / into the body). The third DOF, humeral rotation, is shown but not highlighted. The image to the right shows a serial linkage system over the Franka Panda arm. In the center is depicted a sphere with a radius equivalent to the length of the upper arm and its center represents the center of the shoulder joint. Depicted in blue are the arcs that the shoulder can trace. The green dotted lines are the predicted linearized motion from a serial linkage design.}
\label{fig: Kinematics}
\vspace{-1mm}
\end{figure}




Two key components of the device are to have a single joint center and maximize coverage of the shoulder's range of motion. The shoulder has 3 degrees of freedom (3 DOF), two controlling the arm positioning in space and the third controlling the axial orientation. The flexion and abduction angles are the two degrees of freedom that dictate the kinematics of the upper portion of the arm.

An example of the kinematic equations for the upper arm's position in space given a sequence of angles is shown below:
\vspace{-3mm}
\begin{align}
    x_e &= rcos(\theta)sin(\phi)\\
    y_e &= rsin(\theta)sin(\phi)\\
    z_e &= rcos(\phi)
\end{align}
\vspace{-1mm}
where $[\theta, \phi, r]$ are the flexion and abduction, and the length of the upper arm, a constant, and $[x_e, y_e, z_e]$ are the position of the elbow in Cartesian space relative to the joint center. These are spherical coordinates, meaning the elbow moves along a perfect arc as depicted in Fig. \ref{fig: Kinematics}.

Most commercial robotic arm designs follow a different design pattern in which a redundant serial linkage system is used. A simplified (3 DOF) version of the kinematics is shown in the equation below:
\vspace{-1mm}
\begin{align}
    x_r &= rcos(\theta_1)(1+cos(\theta_2)+cos(\theta_2+\theta_3)) \\
    y_r &= rsin(\theta_1)(1+cos(\theta_2)+cos(\theta_2+\theta_3)) \\
    z_r &= rsin(\theta_2+\theta_3)+rsin(\theta_2)
\end{align}

where $\theta_i, \; i = 1,2...$ are the robot's joint angles and $[x_r, y_r, z_r]$ is the end effector position relative to the base frame. With enough redundancy, to prevent singularities, the end effector path could remain close to a spherical path with traditional Jacobian linearization. However, the differences in the kinematics are significant and would cause concerns about damaging actuators not designed for backdrivability. Therefore the design must use a single joint center to best model the shoulder joint and use actuators capable of backdriving.



\subsection{Modeling Safety: Reach Cone Constraints for Biomechanically Accurate Joint Limits}
\label{sec: Reach Cone Creation}
\vspace{-1mm}

\begin{table}[t]
\vspace{2mm}
\centering
\caption{Shoulder Joint Limits} 
\label{tab:human_parameters}
\begin{tabular}{cccccc}
    \toprule
    \textbf{Direction} & Flex / Ext & Abd / Add & Med/Lat Rotation \\
    \midrule
    \textbf{Angle} &$[160\textdegree, 49\textdegree]$ & $[174\textdegree, 0\textdegree]$ & $[63\textdegree, 92\textdegree]$ \\
    \toprule
    \end{tabular}\\
    Ranges are from individuals aged 25-39 years old. These measurements are an average of goniometer measurements from a neutral position\cite{reese2016joint}.
\vspace{-1mm}
\end{table}

Spherical joints pose a challenge in defining joint limits as they can not be defined with box limits. Box limits are limits defined as linearly independent $[min, max]$ pairs. These are used for serial linkage manipulators, but cannot describe complex joint limits in which the joint limits are interdependent on the current configuration. The shoulder joint limits would be written as such:
\vspace{-.7mm}
\begin{equation} 
\label{const1}
\begin{split}
\Gamma_{free}  = [\gamma_{min}, \; \gamma_{max}] = f_{\gamma}(\theta, \phi)\\
\Theta_{free} = [\theta_{min}, \; \theta_{max}] = f_{\theta}(\phi, \gamma)\\
\Phi_{free}  = [\phi_{min}, \; \phi_{max}] = f_{\phi}(\theta, \gamma)
\end{split}
\end{equation}

where $\Gamma_{free}$, $\Theta_{free}$, $\Phi_{free}$ are the joint ranges of motion and $\gamma$, $\theta$, $\phi$ are the current configuration: humeral rotation, flexion angle, and abduction angle respectively. 

Due to the asymmetry of the humeral head, it is non-trivial to produce an analytical solution for the coupling between the ranges of motion. Instead, most implementations use large data models. In this paper, joint limits are calculated using the reach cone method to minimize memory usage. In applying this method, the coupling function for humeral rotation is assumed to be constant. 






\begin{figure}[t]
\vspace{2mm}
\centerline{\includegraphics[width=80mm]{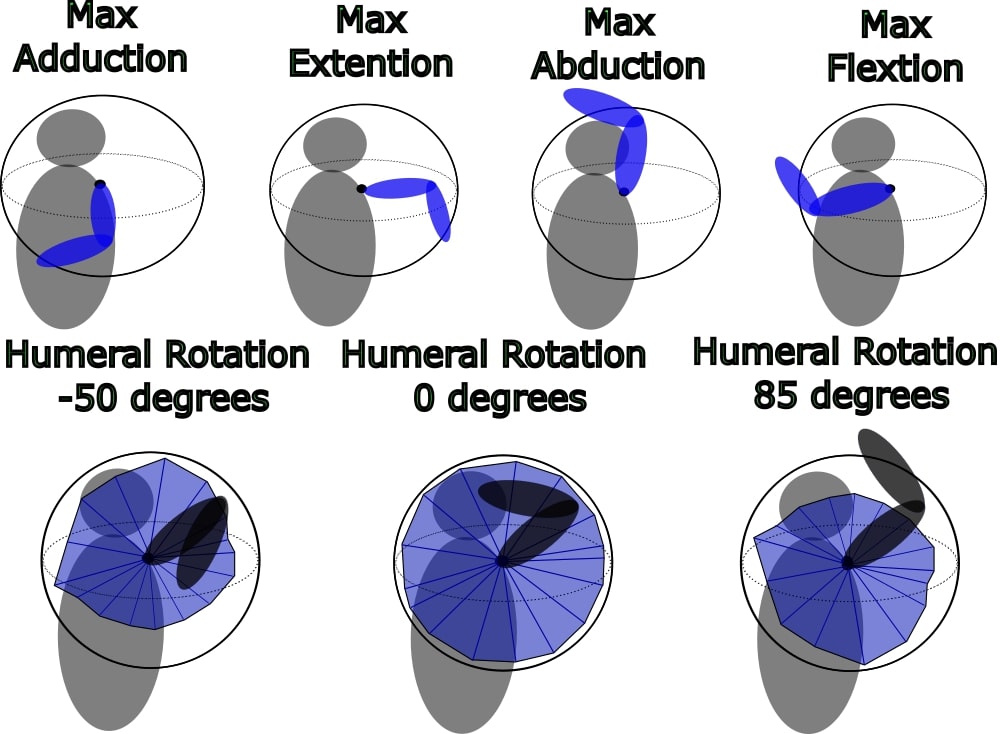}}
\caption{A reach cone starts as 4 angles defining the maximum ROM in each direction. These 4 angles shown in the top 4 images define points on the unit sphere: $[p_1, p_2, p_3, p_4] \in \boldsymbol{P}$, where $\boldsymbol{P}$ is the set of points on the unit sphere defining the joint limits. The second row of images shows different reach cones at different humeral orientations}
\label{fig:SimpleCone}
\vspace{-1mm}
\end{figure}

\subsubsection{Defining the Joint Limits as a Reach Cone}
The reach cone is a discretized set of vectors defining the acceptable orientation region. Data of human shoulder range of motion data is typically gathered via goniometer data which only defines 4 angles, maximum range of motion for flexion, extension, abduction, and adduction, shown in the top row of Fig. \ref{fig:SimpleCone}. Average maxima can be found in Table \ref{tab:human_parameters} \cite{reese2016joint}. To establish a cone shape the data was interpolated to create a total of 64 unit vectors or points on the unit sphere, shown in the second row of Fig. \ref{fig:SimpleCone} where each cone is shown with a different humeral orientation affecting the cone shape.

\begin{figure*}[!ht]
\vspace{2mm}
\centerline{\includegraphics[width=175mm]{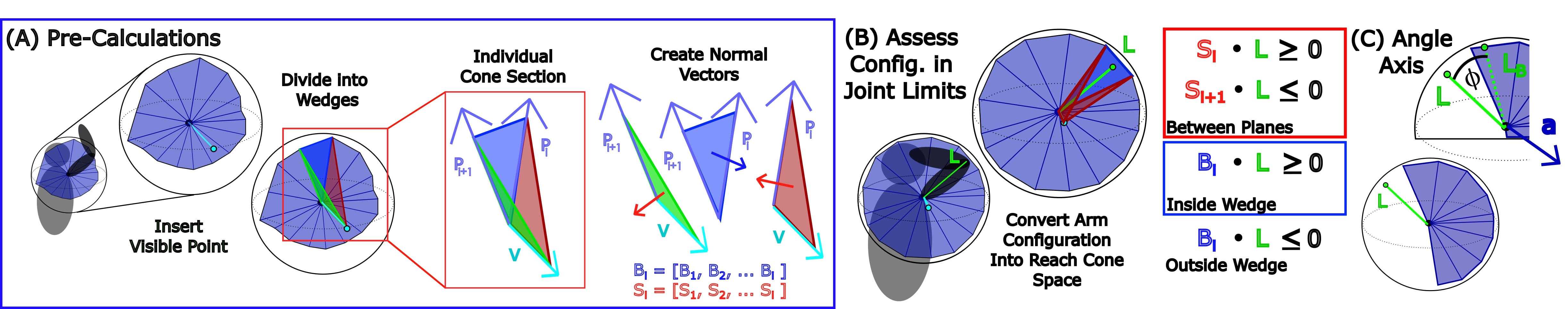}}
\caption{ In part (A) "Pre-Calculations the figure shows the addition of the visible, $V$, in cyan placed in the reach cone to designate the internal region. The visible point is then used to divide the internal space into wedges one of which is depicted in to its three surfaces of red, blue, and green. These surfaces are saved as surface normals $B_i$ and $S_i$ depicted in blue and red. Section (B) shows the calculation to define whether a shoulder orientation belongs in the reach cone. Section (C) shows the final analysis needed for when an orientation is found outside the cone and needs to be pushed back towards the cone in the appropriate direction.}
\label{fig: Complex Cone}
\vspace{-1mm}
\end{figure*}




\subsection{Haptics}
\label{sec: haptics}

A fundamental concern when moving humans is safety. Ideally robotic assistance can use sensors to feel for viscoelastic changes as joints reach their end ranges of motion. To capture both the complex joint limits and multiple persons with highly variable joint limits the SHULDRD was designed with haptics which render a flexible virtual environment. 

The haptic algorithm includes a free-ROM (range of motion), $\Theta_{free}$, and near-end-ROM. When the device is pushed in the free-ROM area, the motors provide viscous damping and transition to providing 3D virtual tendon stretching resistance as the shoulder nears its end range of motion. This simulates the transition as one would stretch the shoulder's tendons before hitting the joint limit, which might cause soft tissue damage. The combined reaction torque is calculated using the following equations:



\begin{equation}
\begin{cases} 
    \boldsymbol{\tau} = (k_p\boldsymbol{\theta}_e - b\boldsymbol{\dot{\theta}}) & \theta \notin \boldsymbol{\Theta}_{free}\\
    \boldsymbol{\tau} = b\boldsymbol{\dot{\theta}} & \theta \in \boldsymbol{\Theta}_{free}
\end{cases}
\end{equation}


Where $\boldsymbol{\tau} \in \mathcal{R}^3$ is the desired torque from each motor, $k_p \in \mathcal{R}$ is the torsional spring constant, $\boldsymbol{\theta}_e \in \mathcal{R}^3$ is the error between the current orientation and joint, $b \in \mathcal{R}$ is the rotational viscous damping coefficient, and $\boldsymbol{\dot{\theta}}$ is the angular velocity of the shoulder.

\subsubsection{Biomechanics of Tendon and Damping Models}


Researchers have found that the mechanical behavior of the long head of the biceps tendon (and most tendons and ligaments in the body) is best described by an initial non-linear, or "toe" region, followed by a linear region as shown in the load vs. displacement curve resulting from a uniaxial tensile Instron test conducted in \cite{KOLZ2015940}. To translate this curve back into joint space it is necessary to convert the translational experimental data to rotational space with the arc length equation relating $S$ the arc length with $r_{ma}$, the moment arm of the supraspinatus muscle about the rotational center of the shoulder in abduction from \cite{kuechle1997shoulder} at 1.54 cm, and $\theta$ the angular displacement.




The translated curve is scaled down to fit the specifications of the device's motors and shown as the Ideal Human Tendon Response curve in Fig. \ref{fig: tendon_data}. A similarly scaled linear torsional spring constant was calculated to be 0.072 Ncm/\textsuperscript{o}.


Moving to a more fitting non-linear model, the torsional spring constant was replaced with the following torsional spring equation derived from the experimental data in \cite{KOLZ2015940}: 
\vspace{-1mm}
\begin{equation}
\label{eqn: tendon 1}
    K_p(\boldsymbol{\theta_e}) = -0.002\boldsymbol{\theta_e}^2 + 0.081\boldsymbol{\theta_e} -0.093
\end{equation}
\begin{equation}
\label{eqn: tendon 2}
    \boldsymbol{\tau} = (K_p(\boldsymbol{\theta_e})\boldsymbol{\theta_e} - b\boldsymbol{\omega})
\end{equation}

Our viscous damping value was approximated from \cite{engin1984damping} which showed a range from $b$ = [0.30, 0.55], and the constant value 0.35 N-m-s/rad was chosen.

\subsubsection{Virtual Tendon Stretch Calculation}
\label{sec: stretch}
As shown in equations \ref{eqn: tendon 1} and \ref{eqn: tendon 2} it is necessary to calculate the angular error or angular distance from the joint limits

Calculating the inclusion of a specific state, $\theta$, within the reach cone efficiently involves a series of steps outlined in \cite{wilhelms2001fast}. The first of which is defining a visible vector or point that is in $ \boldsymbol{\Theta}_{free}$ and defines which half of the sphere shown in Fig. \ref{fig: Complex Cone} (A) is the reachable space. Next is calculating all the surface normals of the wedges that comprise the reach cone:
\begin{align}
    \boldsymbol{B}_i = \boldsymbol{P}_i \times \boldsymbol{P}_{i+1}\\
    \boldsymbol{S}_i = \boldsymbol{V} \times \boldsymbol{P}_{i}
\end{align}

where $\boldsymbol{P}_{i} \in \mathcal{R}^3$ is the unit vector formed from the joint center to the point defined as part of the edge of the reach cone, $\boldsymbol{B}_{i} \in \mathcal{R}^3$ is the surface normal of the edge triangle with the positive direction pointing into the cone, $\boldsymbol{S}_{i} \in \mathcal{R}^3$ is the surface normal of the internal surfaces subdividing the cone and $\boldsymbol{V} \in \mathcal{R}^3$ is the visible point vector, the vector within the reach cone used to help subdivide the space. 

Fig. \ref{fig: Complex Cone} (A) next shows the surfaces that are formed from the $\boldsymbol{P}_{i}$, $\boldsymbol{P}_{i+1}$ and $\boldsymbol{V}$ vectors. The $\boldsymbol{P}_{i}$ vectors are shown in purple and the visible vector is in cyan. They create the 3 surfaces shown in red, green, and blue whose normals are, $\boldsymbol{B}_{i}$, blue, and $\boldsymbol{S}_{i}$, red.

Next is the algorithm to determine the arm's placement in space relative to the joint limits seen in Fig. \ref{fig: Complex Cone} (B). The first step in the algorithm determines which internal surfaces the arm lies between. A vector is created from the flexion and abduction angles of the arm. The raw values from the motors are intrinsic angles and the clinically relevant angles are extrinsic angles. Normally a Jacobian, transform is applied, however, this design uses a single joint center which makes the Jacobian a rotation matrix and exact, with no linearization needed as shown in \ref{sec: kinematics}. This rotation is applied to convert the motor angles to human joint angles used in the reach cone space. The transformed angles are then used to form the vector $\boldsymbol{L}$. A dot product is then taken with each $\boldsymbol{S}_{i}$ until the conditions describing its containment between consecutive planes are met. This condition is shown below and in \ref{fig: Complex Cone} (B) boxed in red.
\vspace{-2mm}
\begin{equation}
    \boldsymbol{S}_i \cdot \boldsymbol{L} \geq 0 \:\: and \:\: \boldsymbol{S}_{i+1} \cdot \boldsymbol{L} \leq 0
\end{equation}
\vspace{-1mm}
The last step is to calculate whether the arm is within the wedge using the outer surface corresponding to this particular wedge, $\boldsymbol{B}_i$. This dot product also produces the linear distance from the outer surface, $d$, as all the surfaces and vectors share a common point which is the joint center. If $d$ is positive the arm is within the reach cone, if the value is negative the arm is outside the reach cone.





To produce the proper virtual forces in the component directions of the motors the distance, $d$, will need to be converted into $\theta_e$ for each of the flexion and abduction directions. To do this $\boldsymbol{L}$ is projected onto $\boldsymbol{B}_i$ and normalized, shown below:
\vspace{-.7mm}
\begin{equation}
\begin{split}
    \boldsymbol{L}_B = proj \rightarrow \boldsymbol{L} \rightarrow \boldsymbol{B}_i = \frac{\boldsymbol{L} - (\boldsymbol{L} \cdot \boldsymbol{B}_i) \boldsymbol{B}_i}{\|\boldsymbol{L} - (\boldsymbol{L} \cdot \boldsymbol{B}_i) \boldsymbol{B}_i\|}
\end{split}
\end{equation}

Next, the angle between the original and projected vector is calculated to get the angle needed to rotate the vector onto the reach cone. The last step is to get the axis about which the vector must be rotated to get the full angle-axis rotation transform. This vector will give the rotation about each orthogonal axis. These equations are shown below:
\vspace{-1mm}
\begin{align}
    \phi = cos^{-1}(\boldsymbol{L} \cdot \boldsymbol{L}_B)\\
    \boldsymbol{a} = \boldsymbol{L} \times \boldsymbol{L}_B
\end{align}

\begin{figure}[t]
\vspace{2mm}
\centerline{\includegraphics[width=\linewidth]{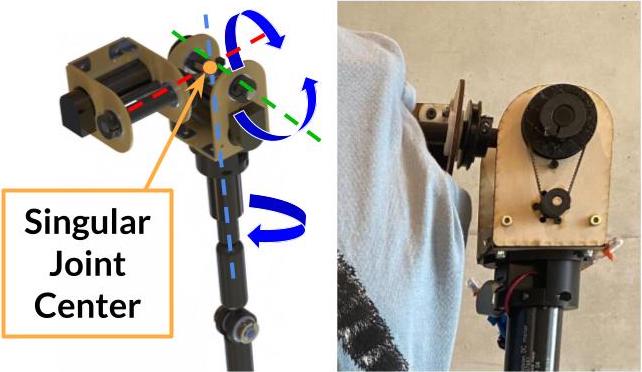}}
\caption{Full SHULDRD device mounted onto a mannequin. The device is oriented to align its zero positions with anatomical zero. The single joint center is highlighted in yellow. The axes depict the 3 axes of rotation: red for flexion, green for abduction, and light blue for humeral rotation. As is seen in the image all 3 axes intersect at the orange joint center.}
\label{fig:fulldevice}
\vspace{-1mm}
\end{figure}

\subsection{Mechanism}

The system referred to as SHULDRD (Shoulder Haptic Universal Limb Dynamic Repositioning Device) is an augmented u-joint with an added degree of freedom. The motors and encoders measure joint angles and render virtual environments. There is also a detachable forearm with a hand. The full design can be seen in Fig. \ref{fig:fulldevice}.


\subsubsection{Dual Joint (flexion and abduction)}
As described in section \ref{sec: kinematics}, it is fundamental that the device has a single center of rotation for the flexion and abduction to guarantee transfer from human kinematics. The Dual joint achieves this through its T-shaped shaft with a single center of rotation like the shoulder.

The Dual Joint has 2 motor-encoder pairs for measurement and rendering. Each motor-encoder pair is housed in a basket comprised of laser-cut and off-the-shelf components. The motors are connected to a belt and pulley system with a 1:3.33 torque ratio.





\subsubsection{Single Joint (humeral rotation)}


The Single Joint is composed of a motor adapter, a motor-encoder pair, and a coupler to the forearm. The motor adapter rigidly attaches the motor to the abduction motor housing which moves as the abductor motor rotates.

\vspace{-1mm}
\subsection{Electronics}


The Arduino Mega 2560 Rev3 is chosen for the embedded system and communicates to a PC for data logging through USB. The power system consists of a separate 24V and 8V rail. The 8V rail powers the Arduino Mega and three SKU-DRI0002 motor controllers. The 24V rail provides power to the maxon 118800 motors. All motors have a HEDL-5540 optical incremental encoder that Arduino powers to supply 5V. Position data from the encoders are logged via a serial monitor.


\begin{figure}[t]
\vspace{2mm}
{\includegraphics[width=.49\linewidth]{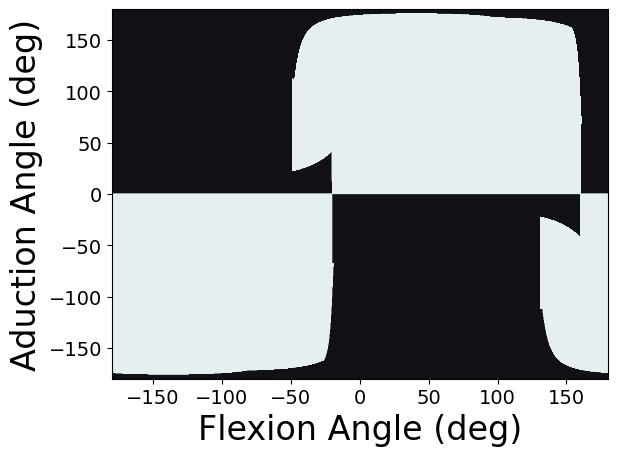}}
{\includegraphics[width=.49\linewidth]{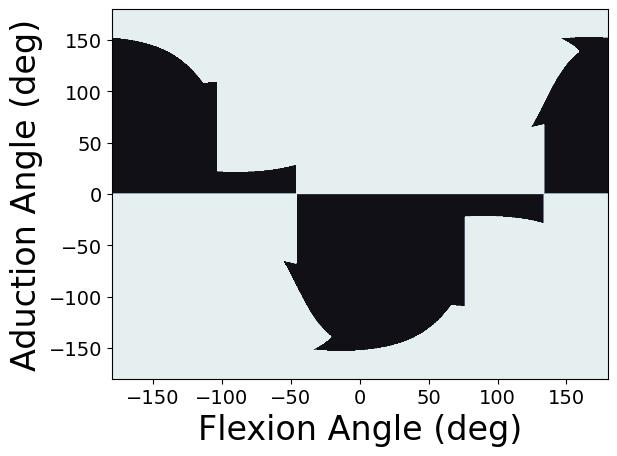}}
\caption{On the left is the C-space of the human shoulder. On the right is the C-space of SHULDRD. White indicates that a configuration is within the reachable space and black indicates it is not. The reachable space of the human shoulder reaches 51.5\% of the spherical space, SHULDRD reaches 71.25\%, the shoulder but not SHULDRD reaches 4\%, and SHULDRD but, not the human shoulder reaches 28.75\%. This result shows a meaningfully larger range of motion from SHULDRD than the human shoulder.}
\label{fig: Shoulder ROM}
\vspace{-1mm}
\end{figure}


\subsection{Software}


The software is based in C++ and the architecture from the Hapkit \cite{martinez2019evolution} was used as the base for rendering the soft tissue forces.
In addition to tactile feedback, the SHULDR device can give audible feedback by augmenting the PWM frequency of the motors into the audible range for humans, to indicate when the shoulder tendons are stretching.






\section{Experiments and Simulations}

\begin{figure*}[t!]
\vspace{2mm}
{\includegraphics[width=34.5mm]{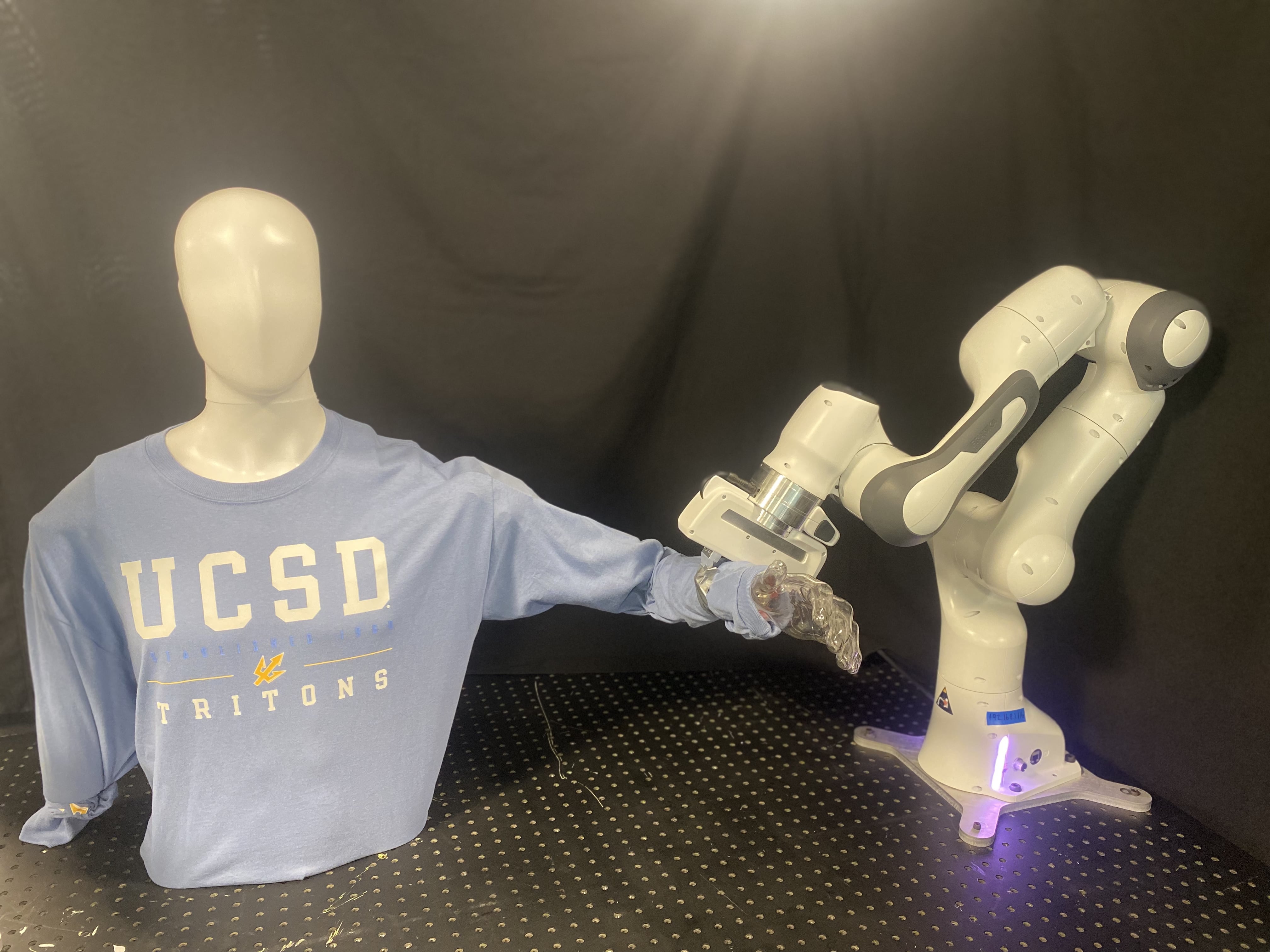}}
{\includegraphics[width=34.5mm]{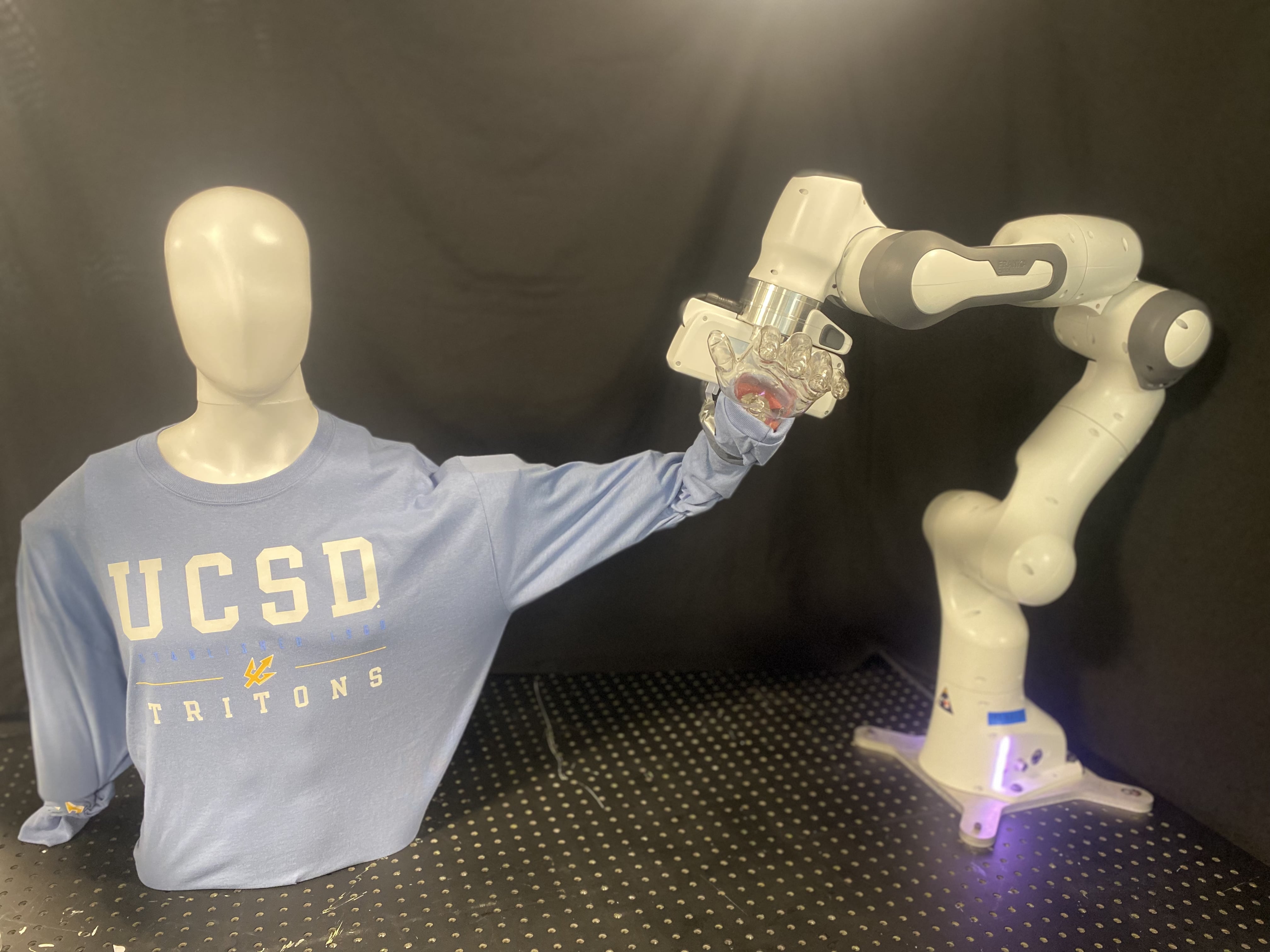}}
{\includegraphics[width=34.5mm]{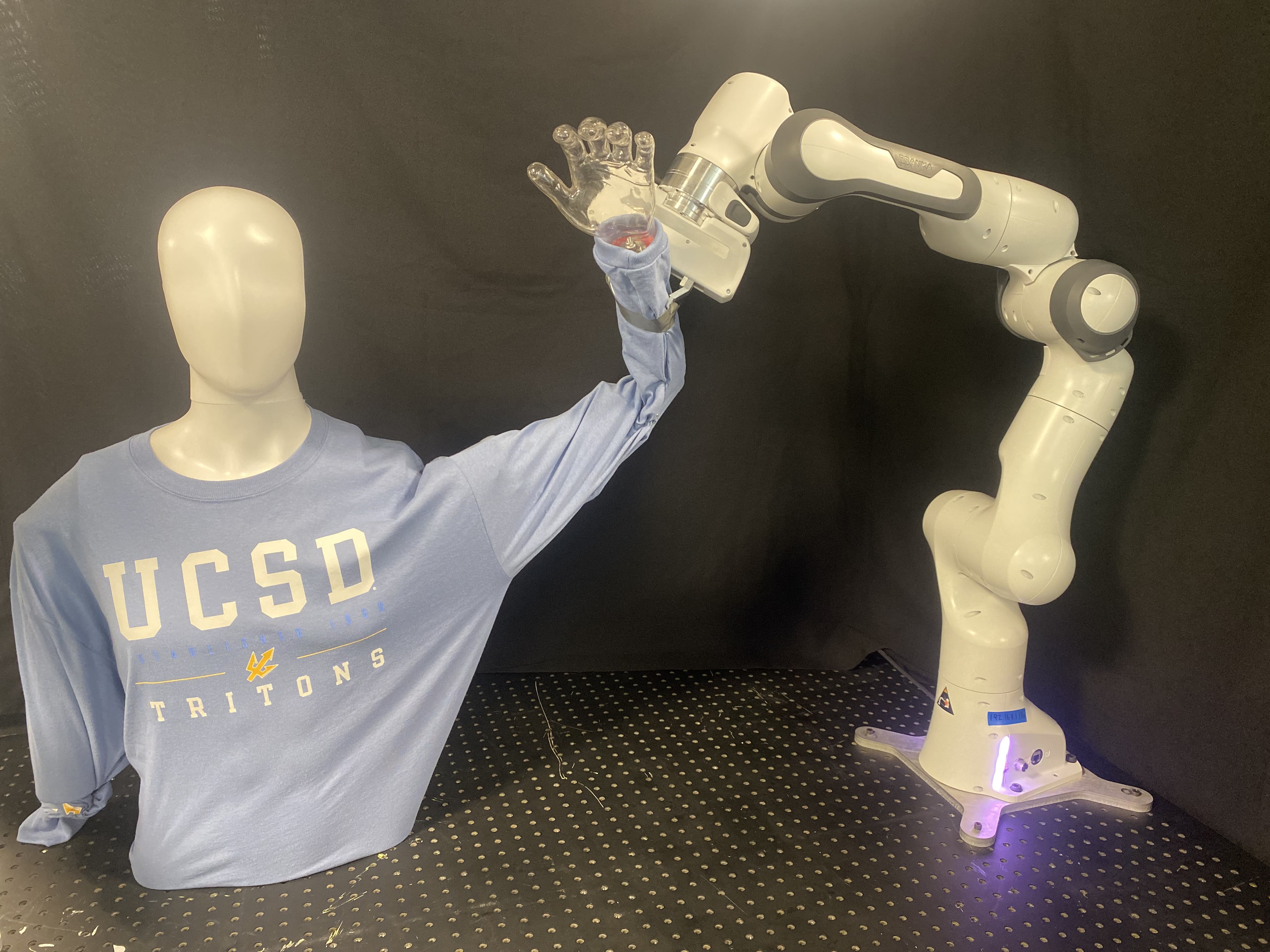}}
{\includegraphics[width=34.5mm]{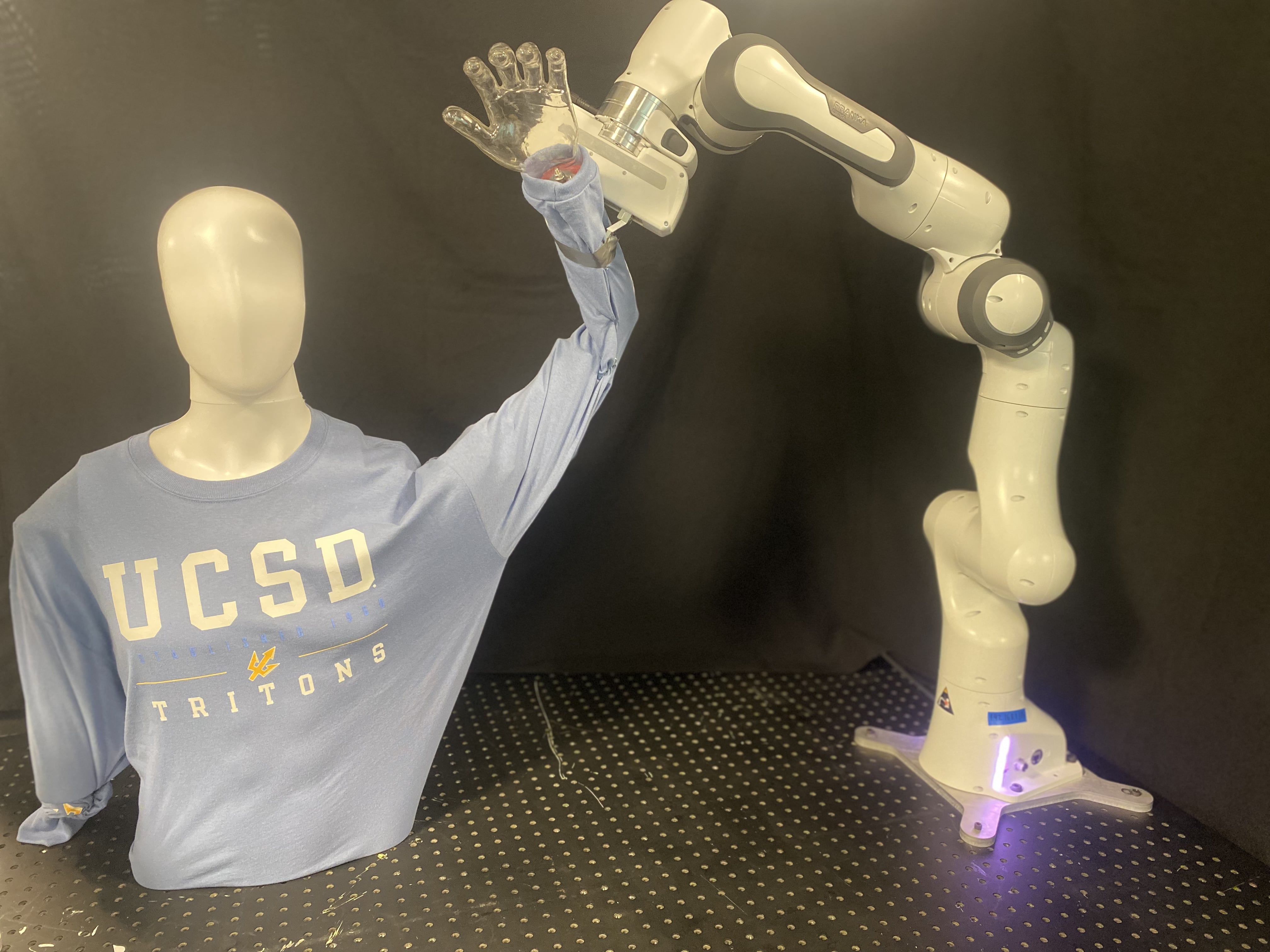}}
{\includegraphics[width=34.5mm]{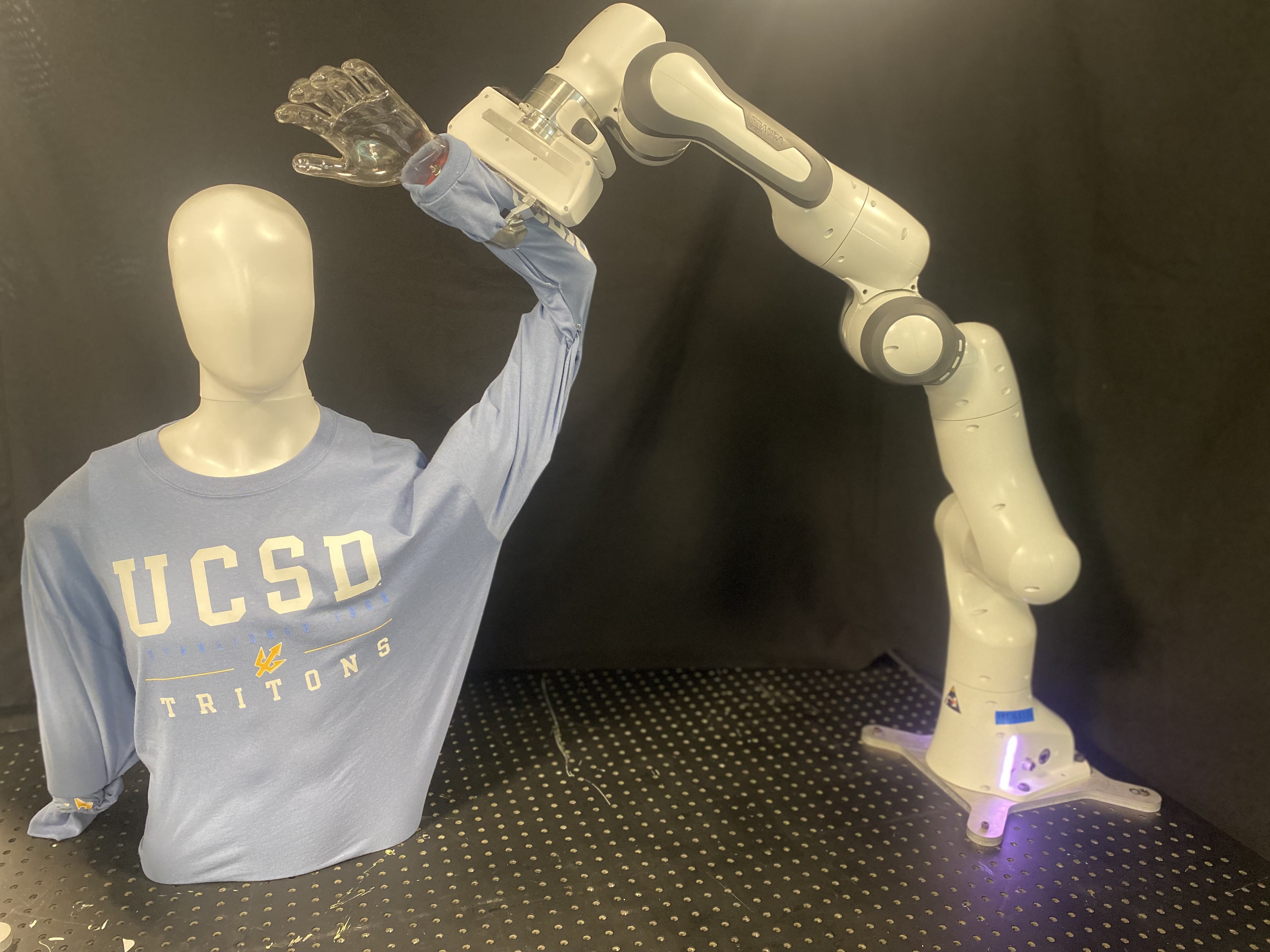}}
\caption{The series of images above show snapshots of a trajectory in which the Franka Panda arm can move the SHULDRD through a significant range of motion limited by the Franka arm's reachable space. In this trajectory the Franka arm has its default gripper holding the arm. In other trajectories, a load cell and three-finger gripper (Right-hand) were used to validate the ability to sense the resistive torques of the motors.}
\label{fig:testshuldrd}
\vspace{-1mm}
\end{figure*}


\subsection{Comparing Reachable Space of Human and Device}
\label{sec: reachable space}

The reachable space of the human shoulder and SHULDRD with configurations that allowed for maximum reach in the four principle directions was taken. The procedure followed that from sections \ref{sec: Reach Cone Creation} and \ref{sec: stretch} and used data from Table \ref{tab:human_parameters}. For the device internal sensors were used to collect data as the device was moved through to its maximum 4 angles.

Fig. \ref{fig: Shoulder ROM} shows the configuration space of the human shoulder and the SHULDRD side-by-side. As the images indicate the SHULDRD exceeds the reachable space of the biological shoulder. The Human shoulder is capable of reaching 51.5\% of the configuration space while the SHULDRD reaches 71.25\%. There is only a 4\% region the human contains that the device cannot, however, there is a 28.75\% region the SHULDRD contains outside of the human shoulder's ROM.

\begin{figure}[t]
\centerline{\includegraphics[width=85mm]{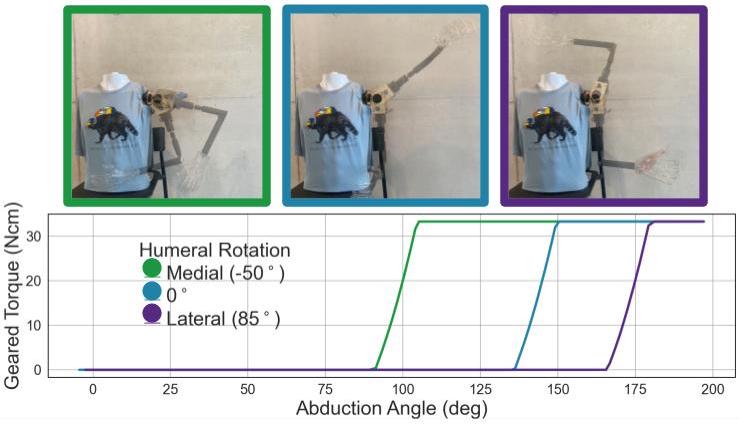}}
\caption{The graph above shows multiple trials moving the SHULDRD with increasing humeral rotation. The motor torque (Ncm) is graphed against the measured angle (deg) from the device and shows how the change in configuration changes the activation angle of the motors which simulates the transition to new joint limits.}
\label{fig: Shoulder humeral example}
\vspace{-1mm}
\end{figure}

\subsection{Humeral Joint Coupling Experiment}

A key biomechanical aspect of the shoulder complex is the coupled nature of the joint limits in each DOF. To validate the SHULDRD’s ability to replicate this behavior, the device was moved in a repeated arc constrained to pure abduction while changing the humeral angle for each trial and observing the onset of motor torque activation. The internal encoders were used to measure the real-time position of both the humeral and abduction rotations and the commanded output torque was recorded. 

As shown in Fig. \ref{fig: Shoulder humeral example} the device activates the motors at different abduction angles dependent on the humeral rotation as the device transitions between joint limits. A -50\textsuperscript{o} internal rotation of the humerus gave the smallest range of motion at 90\textsuperscript{o}. An external rotation of 80\textsuperscript{o} gave the largest range at 165\textsuperscript{o}.



\subsection{Rendering Biomechanically Accurate Tendon Forces}

To ensure the accuracy of the human tendon characteristics within the haptic environment an experiment was conducted using a 1 DOF robotic system. This simplistic system senses the transition from the free ROM into virtual tendon stretching. The testing setup included a Mark-10 Series TT03 torque gauge which captured real-time data from the motor outputting first a torsional spring constant of 0.072 Ncm/\textsuperscript{o} then a non-linear spring equation which ranges from [0 Ncm/\textsuperscript{o}, 0.243 Ncm/\textsuperscript{o}].

\begin{figure}[t]
\centerline{\includegraphics[width=80mm]{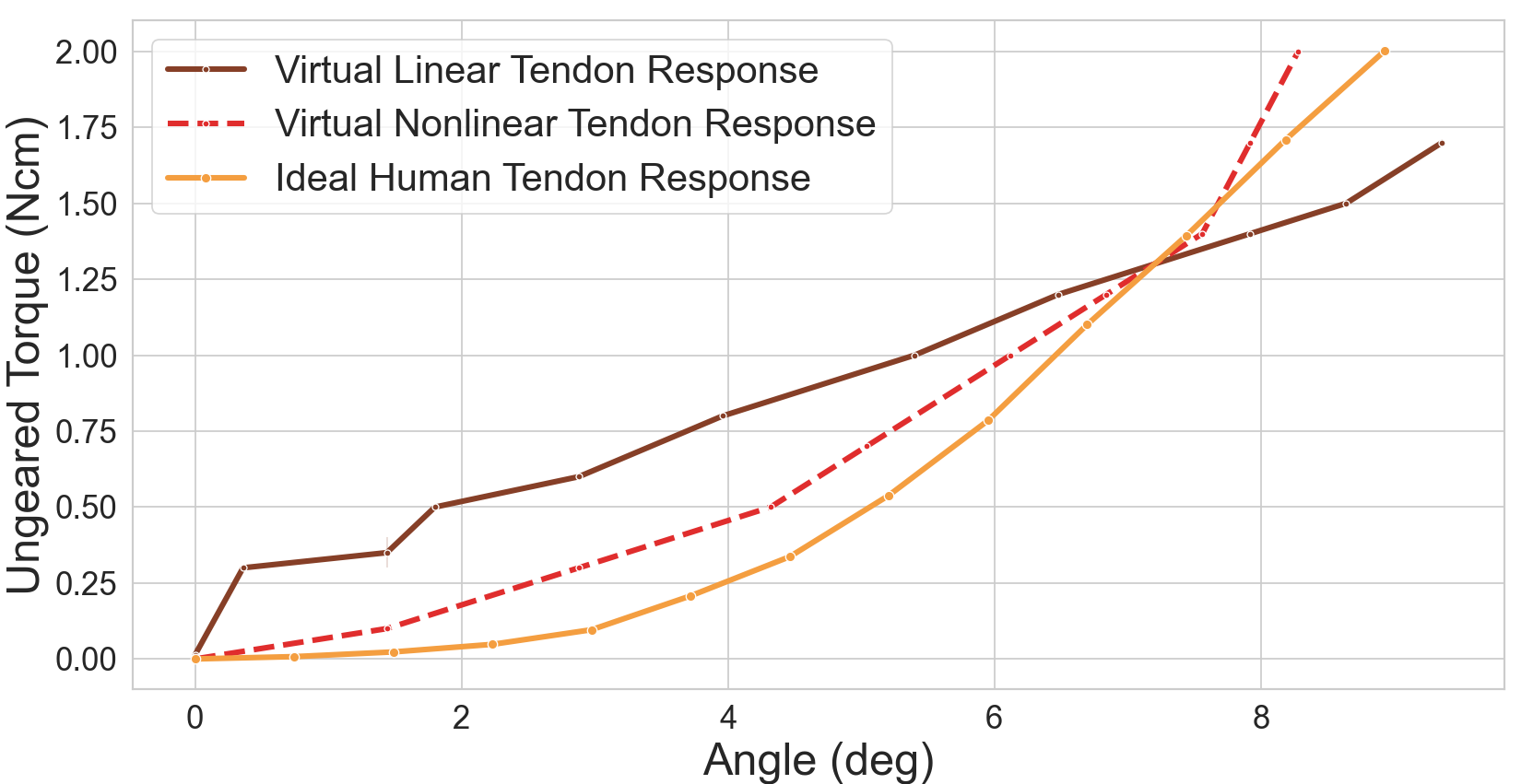}}
\caption{The orange line represents the loading curve from Instron tensile tests of the LHB tendon \cite{KOLZ2015940}, showing an initial non-linear elongation phase followed by a linear phase before yielding. The maroon line shows the test results when the curve is approximated using a torsional spring constant of 0.072 Ncm/°, providing a linear response. Lastly, the red line reflects the approximation of the human tendon response used in the experimental setup.}
\label{fig: tendon_data}
\end{figure}

The data is plotted in Fig. \ref{fig: tendon_data} alongside the ideal tendon curve. The root mean square errors for the linear and non-linear tendon models were 0.2401 and 0.0795, respectively. Comparing both experimental tendon responses, it is clear that the non-linear model closely followed the behavior of the ideal human tendon response. These results indicate that the SHULDRD is capable of accurately rendering human tendon forces.

\subsection{pHRI: using the SHULDRD as a human subject}

The Franka Panda robotic manipulator, ATI Axia80 M8 transducer force-torque sensor, and 2 different grippers were used to manipulate the SHULDRD. The manipulator was positioned to grasp the arm and given a waypoint trajectory to move the device. The path was chosen to activate the virtual tendons and return the free range of motion. The time series images for one of the trajectories, with the Franka and native gripper, can be seen in Fig. \ref{fig:testshuldrd}. The force-torque sensor when integrated was capable of sensing the resistive torques of the virtual joint limits, in a manipulation task. This comparison data of passive and active motors is shown in Fig. \ref{fig: results}. 






\section{Discussion and Conclusions}

In this paper, the device replicates the essential parts of human shoulder motion and the limitations necessary for real-world planning and learning algorithms meant to enhance safety for human-robot physical interaction. The SHULDRD shows the ability to measure its angular position and provide accurate complex joint constraints, tendon forces, and damping which mimic the biological constraints and parameters.

Some inaccuracies in the modeling include the magnitude of the forces which were chosen at a much lower maximum torque for user safety and use in with smaller more affordable manipulator platforms. Larger motors and more expensive motor drivers could replace the current electronics to increase sensing capabilities and output torque.

As with all robotics systems, it is important to understand where singularities might exist. In the SHULDRD, the minimum set of actuators was chosen to cover 3 degrees of freedom. While this afforded the device simplicity and exact transforms, there is a singularity at 90\textsuperscript{o} abduction when the humeral rotation and flexion axes are aligned which inhibits some trajectories.

Future applications of this work include testing path-planning algorithms to help with self-feeding, dressing, search and rescue, and telemedicine for safety without the risk to human subjects. This should increase the ability to test learning algorithms that require large amounts of data and trials to run properly and test algorithms that do safe exploring near the end range of motion for the shoulder.

\clearpage
\bibliographystyle{ieeetr}
\bibliography{refs}

\end{document}